\documentclass[10pt, a4paper]{article}

\usepackage[final]{lrec2026} % this is the new style
\usepackage{amssymb}
\usepackage{amsmath}
\usepackage{amsthm}
\usepackage{transparent}
\usepackage{tikz}
\usepackage{csquotes}
\usepackage{multirow}
\usepackage{pgfplots}
\pgfplotsset{compat=1.18}
\usetikzlibrary{external,positioning,shapes,arrows, calc, fit}
\usepackage{booktabs}

\title{ATLAS: Article Tracking, Linking, and Analysis of Swedish Encyclopedias}

\name{Albin Andersson, Salam Jonasson, Fredrik Wastring, Pierre Nugues} 

\address{Lund University \\
         Lund, Sweden \\
         al3146an-s@student.lu.se,
         ab5015mu-s@student.lu.se,\\
         fr7658wa-s@student.lu.se,
         pierre.nugues@cs.lth.se\\\\
         \textit{Paper originally published in the Proceedings of the LREC 2026}}

\abstract{
The digitization of old encyclopedias represents an important step to improve access to historically structured knowledge. Often, however, this process does not go beyond an optical character recognition, leaving all the underlying structure unexploited. In addition, many encyclopedias had multiple editions reflecting the evolution of knowledge. The lack of structure in the raw text makes it difficult to track changes across these editions.
In this work, we built a pipeline to restore the text structure, where we extract the headwords and identify entries; categorize the entities; match entries across editions; and link entries to a Wikidata item.
We applied this pipeline to the four major editions of \textit{Nordisk familjebok}, an authoritative Swedish encyclopedia published between 1876 and 1951.
We could extract the headwords with an F1 score of 97.8\% and we obtained an F1 score of 93.4\% on the headword classification. On a small-scale evaluation, we reached a 93\% precision on the cross-edition matching, 85\% precision and 16.5\% recall on the Wikidata linking.
This shows that an automated approach to digitized historical knowledge is possible. This should facilitate the preservation of general knowledge and the understanding of knowledge transmission. The datasets and programs are available online.
 \\ \newline  }
 
\begin{document}

\maketitleabstract
\section{Introduction}
Old encyclopedias are valuable pieces of historical knowledge, reflecting the life and ideas of their time. However, much of this knowledge remains locked in unstructured text, making it difficult to analyze it systematically and draw usable conclusions. 

\textit{Nordisk familjebok} is the most comprehensive Swedish encyclopedia of its time. It holds an important place in Swedish literature and used to occupy a prominent place in many Swedish home libraries \citep{ne-nordiskfamiljebok}. \textit{Nordisk familjebok} was published between 1876 and 1951 and had four major editions. The first one (E1) comprises 20 volumes (1876-1899), the second one (E2), 38 volumes (1904-1926), the third one (E3), 23 volumes (1923-1937), and the fourth one (E4), 22 volumes (1951). The second edition still has a high cultural status and is often referred to as \textit{Uggleupplagan}, the `Owl edition' \citep{nordisk-familjebok}. 

These four editions span different periods, reflecting evolving perspectives and knowledge. As a concentrate of their time, they are key sources for the study of intellectual history. Nonetheless, although these editions have been digitized, they still suffer from varying levels of optical character recognition (OCR) quality and inconsistent segmentation markup. This makes them difficult to navigate their content, study the evolution of entries, and relate them to current knowledge.

\begin{figure*}[t]
\centering
\includegraphics[width=\textwidth]{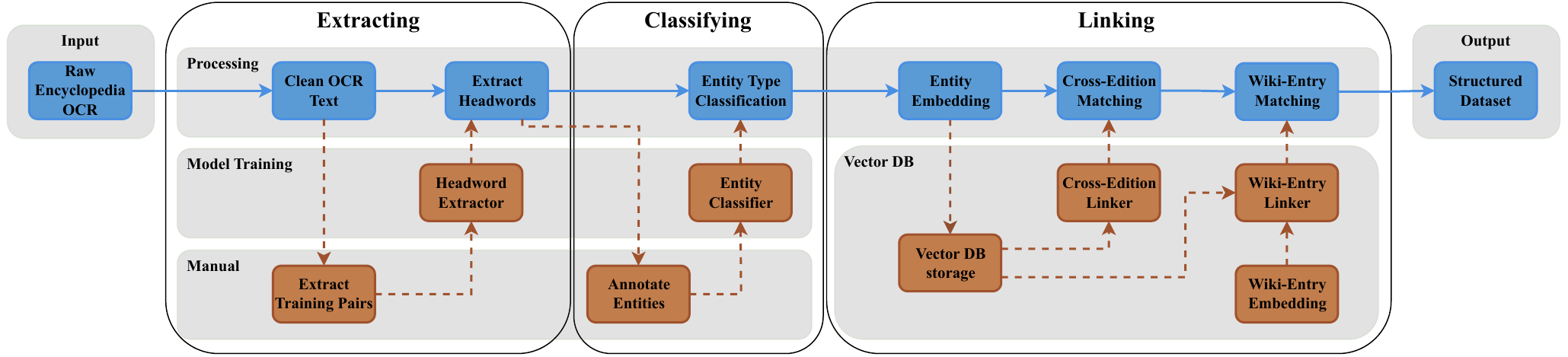}
\caption{An overview of the ATLAS pipeline.}
\label{fig:ATLAS}
\end{figure*}

In this paper, we present ATLAS (Article Tracking, Linking, and Analysis of Swedish encyclopedias), a pipeline for processing historical encyclopedic content. ATLAS aims to manage tasks such as text segmentation into entries and extraction of their headwords; classification of entities into three categories, \textit{Location}, \textit{Person}, and \textit{Other}; matching of a same entry with different versions across editions; and finally linking entries to Wikidata items. Figure~\ref{fig:ATLAS} shows an overview of the ATLAS pipeline.

The contributions of this work are the following:
\begin{enumerate}
    \item We scraped the four editions of the encyclopedia. We preprocessed this dataset and cleaned it to get rid of irrelevant content;
    \item We created a dataset of segmented entries annotated with their headword. We used it to train models to extract the headwords;
    \item We annotated a second dataset of 6000 entries with entity classes and we trained a  classifier; 
    \item We matched entries across editions. We compared each entry of a given edition to all other editions using a sentence embedder;
    \item We linked Wikidata items that had a reference to the encyclopedia entries using the same approach as in the previous step.
\end{enumerate}

The datasets and code required to reproduce the experiments are publicly available on Hugging Face at \url{https://huggingface.co/albinandersson/datasets} and GitHub at \url{https://github.com/SalamSki/EDAN70}.

\section{Previous Work}
Our system builds on work in three main areas: digitization of historical texts, recognition of named entities (NER) in historical documents, and linking of historical encyclopedias to Wikidata.

\subsection{Digitization of Historical Texts}
There are now scores of book digitization projects. Project Runeberg has digitized numerous Nordic texts with an OCR and volunteer proofreading \citeplanguageresource{project-runeberg}. Project Runeberg digitized the four editions of \textit{Nordisk familjebok} and made them accessible online.\footnote{\url{https://runeberg.org/nf/}} 

While the digitization of \textit{Nordisk familjebok} is highly valuable, the OCR quality varies across volumes, and the segmentation into distinct entries remains incomplete. The proofreading is still in progress with varying degrees of completion depending on the edition. This underscores the need for post-OCR correction methods, such as character sequence-to-sequence models proposed by \citet{Ramirez-Orta_Xamena_Maguitman_Milios_Soto_2022}.

\subsection{Entity Recognition}
Named entity recognition (NER) has a long history. We can reformulate it as a sequence annotation problem, where we annotate each word to mark it as a part of a named entity or not. As significant milestones, \citet{Collobert2011} applied a unified feed-forward neural network architecture associated with embeddings. \citet{lample-etal-2016-neural} used long short-term memory (LSTM) networks. Finally, \citet{Devlin2019} applied transformer encoders that improved the feed-forward and LSTM scores. 

Current NER systems build on embeddings or models often pretrained on contemporary text. Historical documents present unique challenges due to evolving language, shifting spelling conventions, and inconsistent formatting \citep{Ehrmann-2023}. \citet{tudor-pettersson-2024-people} showed promising results by fine-tuning pretrained models on historical Swedish text to better handle style variations.

\subsection{Linking Encyclopedias to Wikidata}
In recent years, entity linking (EL) has advanced significantly, driven by improvements in language models and dense vector embeddings. \citet{wu-etal-2020-scalable} introduced a zero-shot approach using dense entity retrieval, while \citet{gillick-etal-2019-learning} demonstrated the effectiveness of dual encoder models for mapping mentions and entities into a shared vector space. Similarly, \citet{logeswaran-etal-2019-zero} developed a zero-shot approach that links textual mentions to a knowledge base without entity-specific training data. These approaches rely on various similarity metrics to compare embeddings. \citet{botha-etal-2020-entity} showed that storing embeddings in vector spaces and performing similarity searches is an efficient strategy for large-scale entity linking. \citet{ayoola-etal-2022-improving, ayoola-etal-2022-refined} proposed methods to incorporate the type in the disambiguation.

Previous projects have also explored linking historical encyclopedias to the Wikidata entity repository. As examples, \citet{nugues-2022-connecting,nugues:2024:main} linked proper nouns in \textit{Petit Larousse illustré}, a French dictionary, and Diderot's \textit{Encyclopédie}. For \textit{Nordisk familjebok} specifically, \citet{ahlin} explored linking location entities for the second edition and \citet{borjesson-etal-2025-matching}  for the first and second editions.

Our work extends these approaches in three key ways: we processed both person and location entities; we covered all four editions of the encyclopedia that were available online before Summer 2025; and we introduced methods for cross-edition entity tracking.

\section{Datasets}
\textit{Nordisk familjebok} is organized as a sequence of entries, where the headwords are ordered alphabetically. To recover this structure, we identified the headwords and we segmented the raw text into entries. We then categorized these entries.

To recognize the entries, a possible solution could be to analyze the image layout of the scans as in \citet{wang-etal-2021-layoutreader} or \citet{Wang2022}. We decided to only use the text and apply a hybrid method instead. 
We first wrote rules that we applied to the corpus to create a ``silver standard'' dataset of annotated headwords. We then extracted a subset of it that we manually corrected to serve as a test set. We trained a model on the headword training set to recognize the headwords and segment the entries. We evaluated it on the headword test set. We created a second annotated dataset of categorized entries to train the entry classification models.

\subsection{Raw Text Collection}
Each edition of \textit{Nordisk familjebok} hosted on Project Runeberg corresponds to a set of URLs organized by volumes following the original division of the encyclopedia. Furthermore, each page has a specific URL component that follows the volume URL. We used this structure to scrape the \textit{Nordisk familjebok} editions.

Using the links, we downloaded the HTML content of each page and extracted the OCRed text. The HTML markup structure is regular and makes it trivial to find the beginning and end of the text. The OCRed content contains additional HTML tags. We observed that bold tags, \verb=<b>=, were used to encapsulate headwords at the beginning of an entry. We kept them and removed all the other tags. 

\subsection{Layout Cues}
A key challenge of the digital version is the inconsistency in the marking of headwords. Although the first two editions, E1 and E2, have been proofread with a generally coherent use of HTML tags to separate headwords, the third and fourth editions, E3 and E4, have no headword marking. This inconsistency makes it impossible to apply a direct headword extraction.

A quick manual check showed that all occurrences of the bold tag in the first two editions are headword encapsulations. However, this markup is not systematic. In addition, very few headwords in the last two editions are marked this way.

The raw text at this stage is then under or oversegmented depending on whether we use the \verb=<b>= markup or the new lines to delimit the entries:
\begin{itemize}
\item The bold tags always correspond to a headword, but many headwords are not marked in the first and second editions, and nearly none in the third and fourth ones. 
\item Entries always start with a new line, but many long entries consist of two or more paragraphs and may include other types of content such as image and figure captions, footnotes, and page numbers. Therefore, we could not use the new lines as an entry boundary marker.
\end{itemize} 

\subsection{Headword Dataset}
\label{sec:headwords}
We created the headword dataset from the headword-annotated entries in E1 and E2. We considered all the paragraphs in the corpus. We identified those starting with the \verb=<b>= tag and we extracted their content using regular expressions.

The structure of the headword dataset consists of two items: The input and the label. As input, we used the raw text of a paragraph from its start and up to 500 characters. As label, we used the extracted headword if we could find one or nothing otherwise. For example, for the \textit{Lund} entry, we have:
\begin{description}
\item[Input:] Lund, uppstad i Malmöhus län\ldots beskaffenhet. I all[mänhet]\\
  ``Lund, a city in Malmöhus County\ldots nature. In gen[eral]''
\item[Label:] Lund
\end{description}
We restricted the paragraphs with no headword to start with a capital letter as with: 
\begin{description}
\item[Input:] Sammanfattningen af dessa nya\ldots kan man anse Brandes, hvilken\\
``The summary of these new\ldots one can consider Brandes, who''
\item[Label:] None
\end{description}
and we discarded the rest.

We obtained 308,448 paragraphs in total, where about 80\% contained a headword. Table~\ref{tab:headword_dataset_res} shows the detailed counts. Some of the entries were duplicates. We removed them and this resulted into 305,675 entries. We split the dataset into a training set of 300,675 entries and a test set of 5,000.

\begin{table}[b]
    
    \centering
    %\resizebox{\columnwidth}{!}{
        \begin{tabular}{l|rrc}
        \hline
            \textbf{Editions} & Positives & Negatives & \textbf{Total} \\ \hline
            First             & 114,770   & 11,920    & 126,690        \\
            Second            & 132,264   & 49,494    & 181,758        \\
            \textbf{Total}    & 247,034   & 61,414     & 308,448   \\
            \hline
        \end{tabular}%
    %}
    \caption{Headword dataset, where the positive paragraphs start with a headword.}
    \label{tab:headword_dataset_res}
\end{table}

The simple \verb=<b>= tag rule missed some entries and headwords.
We manually curated the test set. Out of its 5000 samples, the \verb=<b>= tag rule had labeled 951 as negatives. Our manual inspection found 320 FNs. As we did not have the means to carry out a manual correction of the training set, we used it unchanged with about 240,000 positive samples and 60,000 negative ones. 

We used this dataset to train a sequence annotator and extract the headwords, possibly none, see Sect.~\ref{sec:headextractor}. We also used it to segment the text into entries, where we defined an entry by the presence of a headword in the paragraph.

The resulting dataset, \textit{The Nordisk Familjebok Headword Extraction Dataset}, is publicly available.\footnote{\url{https://huggingface.co/datasets/albinandersson/nf-headword-extraction}} See \citetlanguageresource{andersson2026headword}.

\subsection{Headword Category Dataset}
\label{sec:catds}
After the recognition of entries, we classified them into three categories: person, location, or other. We created a second dataset of 6000 entries annotated with these categories to train a second classifier. Given an entry, we assumed that the category of the entity corresponded to the category of its headword. We carried out the annotation in two steps. We started with a first automatic annotation of the headwords. We then manually checked these annotations. 

For the first annotation, we applied an NER model to the entry text in a zero-shot setting. As model, we used KB/bert-base-swedish-cased-ner pretrained on Swedish texts and fined-tuned for sequence annotation tasks on the manually annotated SUC 3.0 corpus \citep{malmsten2020}. The model uses the tags \verb=PRS= for persons, \verb=LOC= for locations, \verb=ORG= for organizations, \verb=TME= for time, \verb=EVN= for events, or nothing if the token is outside these classes. We applied this model to randomly selected entries. As input, we truncated the entry text to the first 500 characters. We used the headword annotation from the model output to determine the entity category.  

When a headword consists of multiple tokens, this method sometimes results in more than one category tag. We applied post-processing rules to merge them into a single label:
\begin{enumerate}
\item If they contain at least one \textit{Location} but no \textit{Person}, then the headword is a \textit{Location}.
\item If they contain at least one \textit{Person} but no \textit{Location}, the headword is a \textit{Person}.
\item If they contain at least one \textit{Person} and one \textit{Location}, this indicates an uncertainty. We classify the headword as \textit{Other}.
\item We default to \textit{Other} if neither \textit{Location} nor \textit{Person} is present.
\end{enumerate}

In the second step, following the zero-shot NER predictions, we extracted a balanced subset of 6000 entries that we verified manually. After manually correcting the annotation, the distribution was slightly altered. We used this labeled dataset to fine-tune a transformer encoder pre-trained on Swedish text for the entry classification task.

As with the headword dataset, we have released this classification dataset online, named \textit{The Nordisk Familjebok Category Classification Dataset}.\footnote{\url{https://huggingface.co/datasets/albinandersson/nf-category-classification}} See \citetlanguageresource{andersson2026classification}.

\section{Method}
The ATLAS system consists of a pipeline of components. Figure~\ref{fig:ATLAS} shows its architecture that we describe now.

\subsection{Headword Extractor and Entry Segmenter}
\label{sec:headextractor}
We modeled the headword extraction task as a sequence annotation task, where each token in an input sentence is classified as either part of the headword (1) or not (0). This enables the model to identify headwords in entries while producing empty masks for the rest of the paragraphs. To segment the entries, we used these results with the rule that an entry always starts with a headword.

We compared two different architectures for this task: A bidirectional LSTM-based model trained from scratch and a fine-tuned transformer encoder trained on Swedish text, KB-BERT \citep{malmsten2020}, with selective layer unfreezing.

\begin{figure}[t]
\centering
\includegraphics[width=\linewidth]{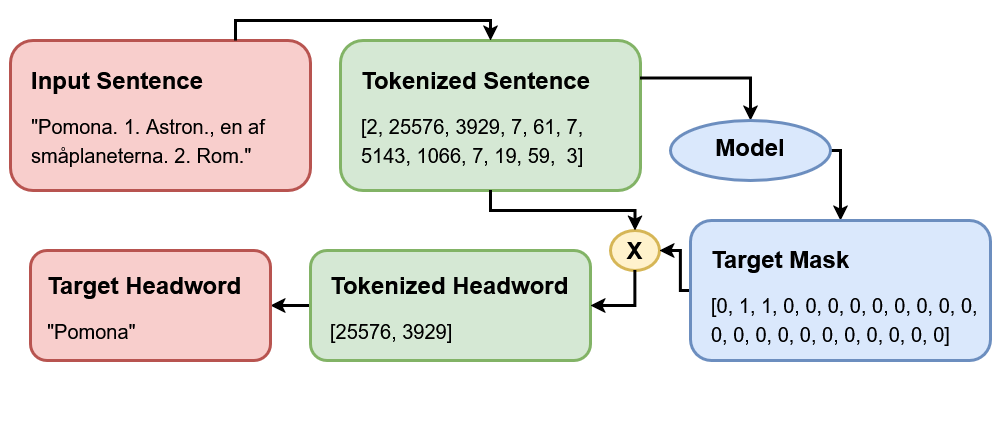}
\caption{The input tokenization and output mask.}
\label{fig:tokenization}
\end{figure}

\paragraph{Data preprocessing.}
For both models, we tokenized the input sentences of the training set and their corresponding headwords using KB-BERT's tokenizer with a maximum sequence length of 100 tokens. We generated the binary target mask by comparing the tokenized sentence with the headword sequences. We annotated the first occurrence of the headword tokens with ones and the rest with zeros, see Figure~\ref{fig:tokenization}.

Additionally, we applied a post-processing to recover the words from the subwords marked with \verb=##= prefixes. In case the annotation of the subwords of a same token was contradictory, positive and negative, we propagated positive classifications backward to all its subwords.

\paragraph{Model architectures.}
The LSTM-based architecture consists of an embedding layer (128-dimensional) followed by a bidirectional LSTM (128-dimensional hidden states). The model processes the embedded sequence and outputs token-level binary classifications.

In the transformer-based approach, we fine-tuned KB-BERT with a task-specific classification head. We employed an unfreezing strategy, where we experimented with different configurations to determine the best number of trainable layers. We applied these models to the raw text of the four editions. 
Both methods enabled us to determine the headwords and segment the text into entries.

\subsection{Headword and Entry Classification}
We then classified the resulting entries into three categories: \textit{Location}, \textit{Person}, or \textit{Other}. We used the pretrained KBLab/bert-base-swedish-cased  transformer encoder that we fine-tuned  on the headword category dataset described in Sect.~\ref{sec:catds}. We added a classification layer with three outputs corresponding to our target types. 

To determine the optimal architecture, we used a systematic layer freezing strategy, where all transformer layers were frozen except for the final $N$ layers. We experimented with $N =$ 0, 2, 4\dots, 12, and selected the best performing model based on the test set performance.

We split the dataset into 70/15/15 train/validation/test sets. We used the AdamW optimizer with a learning rate of $2 \times 10^{-5}$, and batch size of 16. We implemented early stopping with a patience of 3 on validation accuracy to prevent overfitting.

\subsection{Cross-edition Matching}
Tracking entities across different editions is not straightforward. While entries referring to the same entity share similar content across editions, their definitions evolve to reflect temporal changes (e.g., adding death dates for persons or updated population statistics for locations). Additionally, headwords alone are insufficient for matching since they often only contain surnames for persons and some terms can refer to either locations or persons (e.g., \textit{Lund} is both a city and surname).

We embedded the entries in dense vector representations with a sentence transformer model pretrained on Swedish text: S-BERT Swedish Cased \citep{rekathati2021introducing}. We assigned each embedding a unique identifier consisting of the edition number and a sequential index. For instance, the identifier \verb=E2_622= corresponds to the statistician Gottfried Achenwall who has entry number 622 in the second edition.

We stored these embeddings in the Qdrant vector database. For each entry, we ranked the most similar embeddings in other editions using a cosine similarity. We filtered the results by edition prefix to ensure one candidate match per edition. We applied a conservative threshold of 0.75 and we removed the matches that were not symmetrical. We also ensured that, in a match, both editions had the same headword. 

\begin{table*}[tb]
\centering
\resizebox{\textwidth}{!}{
\begin{tabular}{llrllllll}
\hline
\textbf{Entry ID}&\textbf{headword}&\textbf{Type}&\textbf{Edition}&\textbf{E1\_match}&\textbf{E2\_match}&\textbf{E3\_match}&\textbf{E4\_match}&\textbf{QID}\\
\hline
E1\_385 &Achenwall &2&E1&--&E2\_622&E3\_416&E4\_473&Q215933\\
E1\_386&Acheron&1&E1&--&E2\_623&E3\_417&E4\_476&--\\
E1\_387&Acherontia&0&E1&--&--&--&--&--\\
E1\_388&Acherusia&1&E1&--&E2\_625&--&--&--\\
\hline
\end{tabular}
}

\caption{Dataset structure with four entries from the first edition: E1\_385 to E1\_388. The columns contain the entry headword, type (0: Other, 1: Location, and 2: Person), the matches we found in the three other editions, if any, and the QID identifier in Wikidata.}
\label{tab:matcheddataset}
\end{table*}

\subsection{Wikidata Linking}
After matching entries across editions, we linked them to corresponding Wikidata items.  To make the task computationally feasible, we started from the observation that many articles in the Swedish Wikipedia reused entries of the \textit{Nordisk familjebok}. We thus restricted our scope to Wikipedia articles that reference \textit{Nordisk familjebok} as their source. We retrieved these articles and their Wikidata items using SPARQL queries that extracted entries containing a ``described in'' property, \verb=P1343=, and the \textit{Nordisk familjebok} QID as object:
\begin{verbatim}
SELECT ?item
WHERE {
  ?item wdt:P1343 wd:Q678259 .
}
\end{verbatim}
We found about 11,550 such items. A quick inspection showed us that many of the corresponding Wikipedia articles reuse almost identically the text in \textit{Nordisk familjebok}.

Similarly to the cross-edition matching, we stored the first 500 characters of these Swedish Wikipedia articles and the \textit{Nordisk familjebok} entries in a vector database. 
For each \textit{Nordisk familjebok} entry, we computed cosine similarity scores against the Wikipedia candidate embeddings. When the similarity exceeded the threshold of 0.75, and the Wikidata label contained the entry headword, we considered it a match and stored the corresponding Wikidata QID.

We stored the results in a table, where, for each entry, we indicated the headword, its type, the corresponding matching entries in the other editions, if we could find any, as well as the QID. See Table~\ref{tab:matcheddataset} for an example.

\section{Results}
We broke down the results of each step in our pipeline, namely scraping, headword extraction, NER classification, cross-edition matching, and Wikidata linking.

\subsection{Headword Extraction}
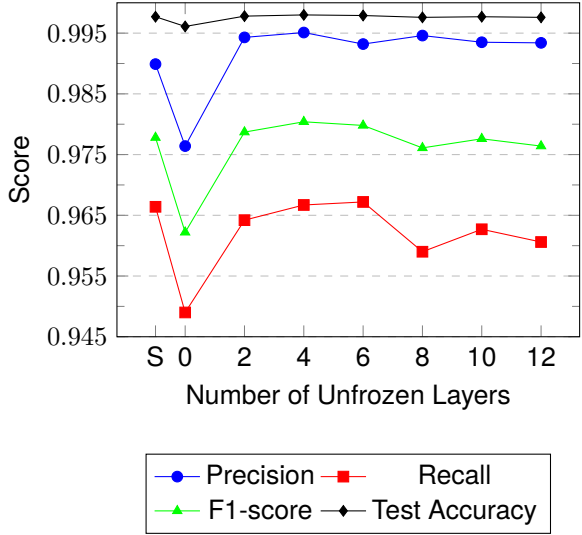
\begin{figure}[tb]
    \centering
    \begin{tikzpicture}
    \begin{axis}[
        width=\columnwidth,
        height=6cm,
        xlabel={Number of Unfrozen Layers},
        ylabel={Score},
        legend style={at={(0.5,-0.35)}, anchor=north, legend columns=2},
        ymajorgrids=true,
        grid style=dashed,
        ymin=0.945,
        ymax=1.0,
        xtick={-1,0,2,4,6,8,10,12},
        xticklabels={S,0,2,4,6,8,10,12},
        ytick={0.945,0.955,0.965,0.975,0.985,0.995},
        minor y tick num=1,
        yticklabel style={/pgf/number format/fixed, /pgf/number format/precision=3},
    ]
    
    \addplot[mark=*, color=blue] coordinates {
        (-1,0.9899) (0,0.9764) (2,0.9943) (4,0.9951) (6,0.9932) (8,0.9946) (10,0.9935) (12,0.9934)
    };
    
    \addplot[mark=square*, color=red] coordinates {
        (-1,0.9664) (0,0.9490) (2,0.9642) (4,0.9667) (6,0.9672) (8,0.9590) (10,0.9627) (12,0.9606)
    };
    
    \addplot[mark=triangle*, color=green] coordinates {
        (-1,0.9778) (0,0.9622) (2,0.9787) (4,0.9804) (6,0.9798) (8,0.9761) (10,0.9776) (12,0.9764)
    };
    
    \addplot[mark=diamond*, color=black] coordinates {
        (-1,0.9977) (0,0.9961) (2,0.9978) (4,0.9980) (6,0.9979) (8,0.9976) (10,0.9977) (12,0.9976)
    };
    
    \legend{Precision,Recall,F1-score,Test Accuracy}
    \end{axis}
    \end{tikzpicture}
        \caption{Performance metrics across different model configurations for headword extraction. On the $x$-axis, ``S'' represents the LSTM model while the figures indicate the number of unfrozen layers in the fine-tuned KB-BERT model.}
    \label{fig:model_comparison}
\end{figure}

We trained and evaluated different model architectures on the headword dataset of Sect.~\ref{sec:headwords}. We compared the LSTM model with various configurations of the fine-tuned KB-BERT model. Figure~\ref{fig:model_comparison} shows the results with different numbers of unfrozen layers. While the fine-tuned model with 4-6 unfrozen layers achieved marginally higher scores, the more lightweight LSTM model had a comparable performance. This minimal performance trade-off combined with the LSTM's lower computational requirements led us to select it for this task. Table~\ref{tab:extractor_performance} shows the results, where we reach an F1 score of 0.9778. Please note that the evaluation is token-based and therefore has a support number of 500,000 values, 100 tokens for each test sample. 

\begin{table}[t]
    \centering
    %\resizebox{0.9\columnwidth}{!}{
        \begin{tabular}{cc|lr}
        \hline
            \multicolumn{2}{c|}{\textbf{Confusion matrix}}                                   & \multicolumn{2}{c}{\textbf{Performance metrics}} \\ \hline
            \multicolumn{1}{c|}{\multirow{2}{*}{486,211}} & \multirow{2}{*}{236}    & Accuracy             & 0.9977           \\
            \multicolumn{1}{l|}{}                         &                         & Precision            & 0.9899           \\ \cline{1-2}
            \multicolumn{1}{r|}{\multirow{2}{*}{905}}     & \multirow{2}{*}{12,648} & Recall               & 0.9664           \\
            \multicolumn{1}{l|}{}                         &                         & F1-score             & 0.9778\\
            \hline
        \end{tabular}%
    %}
    \caption{LSTM score on the manually validated test set with 5000 entries (and 500,000 tokens).}
    \label{tab:extractor_performance}

\end{table}

\subsection{Entry segmentation}
Scraping the four editions into text files and splitting them by new lines results in more than 550,000 potential entries, as shown in Table~\ref{tab:extracted_entires_res}. The  column Scraped shows that the first and fourth editions are fairly similar in size, with the largest edition being the second and most popular. The incomplete digitization of the third edition before Summer 2025 is also reflected by the smaller number of entries. 

Applying the LSTM headword extractor on the scraped files results in 418,221 headword classified entries. This means that approximately 75\% of all paragraphs correspond to the first paragraph of an entry. Table~\ref{tab:extracted_entires_res} also shows the disparity in extraction percentage between the first and second half of the encyclopedia, resulting in more paragraphs per entry for the third and fourth editions. The reason might be that the headword extractor is trained on the first and second editions.

\begin{table}[t]
    \centering
    \resizebox{\columnwidth}{!}{
        \begin{tabular}{c|rrrcr}
        \hline
            \textbf{Editions} & Scraped                      & \multicolumn{2}{|c|}{Extracted}       & \multicolumn{2}{c}{Discarded} \\ \hline
            E1                & \multicolumn{1}{c|}{133,857} & 117,473 & \multicolumn{1}{r|}{0.88} & 16,384          & 0.12        \\
            E2                & \multicolumn{1}{c|}{247,563} & 185,063 & \multicolumn{1}{r|}{0.75} & 62,500          & 0.25        \\
            E3                & \multicolumn{1}{c|}{43,003}  & 26,464  & \multicolumn{1}{r|}{0.62} & 16,539          & 0.38        \\
            E4                & \multicolumn{1}{c|}{131,530} & 89,221  & \multicolumn{1}{r|}{0.68} & 42,309          & 0.32        \\ \cline{1-6} 
            \textbf{Total}    & \multicolumn{1}{c|}{555,953} & 418,221 & \multicolumn{1}{r|}{\textbf{0.75}} & 137,732    & \textbf{0.25} \\
        \hline
        \end{tabular}%
    }
    \caption{Scraped paragraphs, extracted entries, and discarded paragraphs for each edition.}
    \label{tab:extracted_entires_res}
\end{table}

\subsection{Entity Classification}
To fine-tune the KB-BERT models for entity classification, we split the headword category dataset of Sect.~\ref{sec:catds} into a 70:15:15\% split, resulting in a 4200:900:900 sample count for the training, validation, and test sets respectively. Table~\ref{tab:NER_preformance} shows the results with varying numbers of unfrozen layers.

\begin{table}[tb]
    \centering
    \resizebox{\columnwidth}{!}{
        \begin{tabular}{c|rrrr}
        \hline
            \textbf{UFL} & \multicolumn{1}{c}{\textbf{Accuracy}} & \multicolumn{1}{c}{\textbf{Precision}} & \multicolumn{1}{c}{\textbf{Recall}} & \textbf{F1}        \\ \hline
            0                & 0.8078                        & 0.8111                         & 0.8074                      & 0.8072          \\
            2                & 0.9256                        & 0.9248                         & 0.9293                      & 0.9261          \\
            4                & 0.9167                        & 0.9152                         & 0.9201                      & 0.9171          \\
            6                & 0.9267                        & 0.9268                         & 0.9282                      & 0.9271          \\
            8                & 0.9322                        & 0.9328                         & 0.9333                      & 0.9328          \\
            10               & 0.9267                        & 0.9282                         & 0.9278                      & 0.9277          \\
            12               & \textbf{0.9333}               & \textbf{0.9344}                & \textbf{0.9359}             & \textbf{0.9339}\\
            \hline
        \end{tabular}
    }
   
 \caption{Classification performance of the KB-Bert fine-tuned models. UFL means unfrozen layers.}
    \label{tab:NER_preformance}
\end{table}

We notice an almost linear steady increase in all metrics as more layers are unfrozen, reaching the peak with an F1 score of 0.9339 at the maximum of 12 unfrozen layers.  Table~\ref{tab:ner_confusion}  shows its confusion matrix. We applied this model to classify the entities of all extracted entries.

\begin{table}[b]
    \centering
    %\resizebox{\columnwidth}{!}{%
        \begin{tabular}{ccccc}
                &   & \multicolumn{3}{c}{\textbf{Predicted}} \\ \cline{3-5} 
                &   \multicolumn{1}{c|}{12 layers} & \multicolumn{1}{|c|}{Other} & \multicolumn{1}{c|}{Location} & Person \\ \cline{2-2}
            \multicolumn{1}{c|}{\parbox[t]{2mm}{\multirow{3}{*}{\rotatebox[origin=c]{90}{\textbf{True}}}}} & Other    & \multicolumn{1}{c|}{\textbf{0.8727}} & \multicolumn{1}{c|}{0.0848}           & 0.0424           \\ \cline{2-5} 
            \multicolumn{1}{c|}{}                               & Location & \multicolumn{1}{c|}{0.0149}            & \multicolumn{1}{c|}{\textbf{0.9731}} & 0.0119            \\ \cline{2-5} 
            \multicolumn{1}{c|}{}                               & Person   & \multicolumn{1}{c|}{0.0213}            & \multicolumn{1}{c|}{0.0170}            & \textbf{0.9617}
        \end{tabular}%
   % }
        \caption{Confusion matrix of the 12-layer model.}
    \label{tab:ner_confusion}
\end{table}

\begin{figure}[t]
    \centering
    \includegraphics[width=\linewidth]{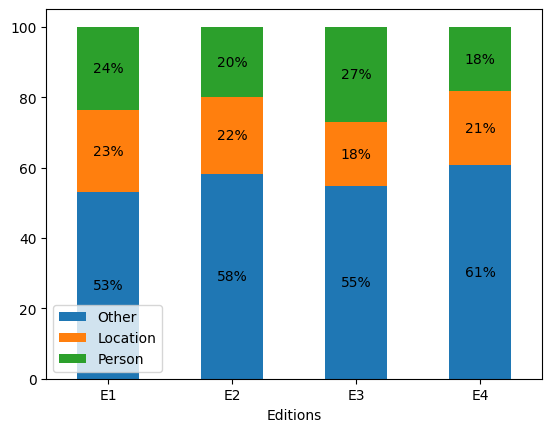}

        \caption{The resulting entity recognition on the extracted entries for each edition.}
    \label{fig:NER_res}
\end{figure}

Table~\ref{tab:categories} shows the category breakdown we obtained. We notice a somewhat general pattern where for each edition, around 50-60\% of the articles are classified as \textit{Other}, 18-23\% as \textit{Locations}, and 20-30\% as \textit{Persons}. Figure~\ref{fig:NER_res} shows the relative proportions for each category per edition.

\begin{table}[t]
    \centering
    \begin{tabular}{lrr}
    \hline
    \textbf{Edition}&\textbf{Locations}&\textbf{Persons}\\
    \hline
    E1& 27,554 &  27,760  \\
    E2& 40,423 &  36,802\\
    E3&  4850 &  7127\\
    E4&  18,794 &   16,216\\
    \hline
    \end{tabular}
    
  \caption{Category breakdown by edition.}
    \label{tab:categories}
\end{table}

\subsection{Cross-edition Matching and Wikidata Linking}
\begin{figure*}[tb]
\centering
\begin{tabular}{cc}
\includegraphics[width=\columnwidth]{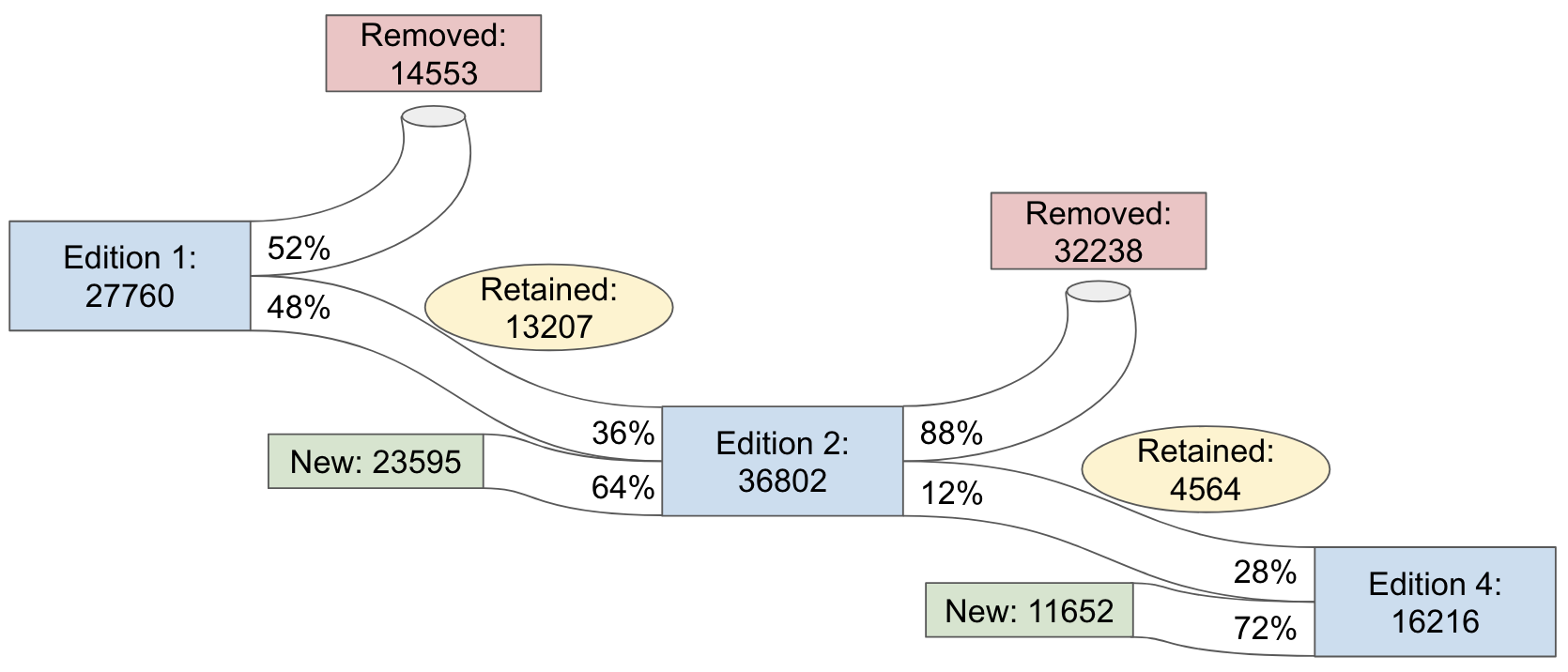}&\includegraphics[width=\columnwidth]{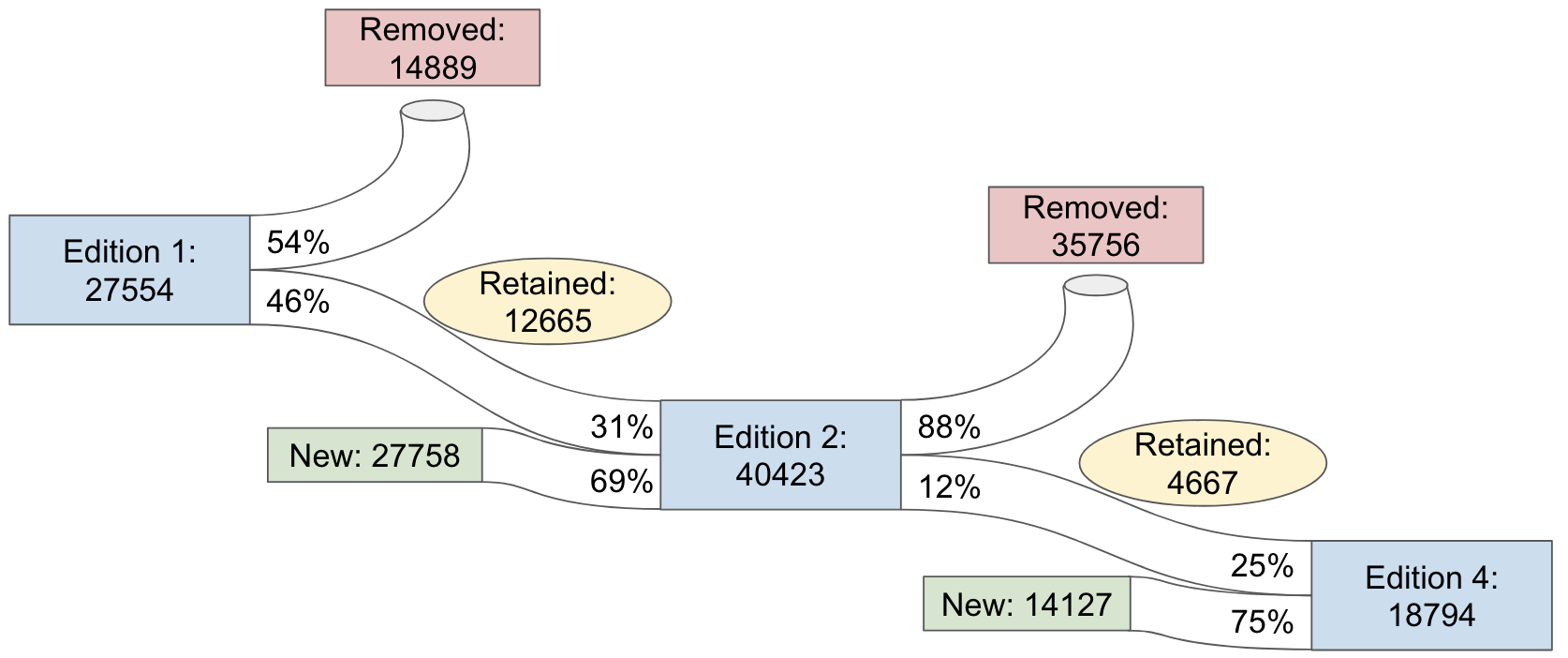}\\
\end{tabular}
    \caption{Additions and removals for person (left) and location (right) entries across editions.}
    \label{fig:matchings_evolution_n}
\end{figure*}

We applied the entry matching method to the four editions. 
Figure~\ref{fig:matchings_evolution_n} shows the number of additions and removals for person and location entries, respectively, in the E1, E2, and E4 editions. We did not include E3 as its digitization was
incomplete before Summer 2025 and hence does not have the same coverage.

The linking process successfully connected 10,964 entries to Wikidata items. The distribution of links varied across the editions: We could create 3766 links to E1 entries, 5088 links to E2, 704 links to E3, and 1406 links to E4. This distribution roughly correlates with the relative sizes of each edition, with E2 being the largest edition overall.

As there is no comparable annotated dataset, we could not extensively evaluate the matching and linking results. We created a limited test set consisting of the person entries with a match in the four editions as in the first row in Table~\ref{tab:matcheddataset}. We can easily verify that two definitions in two different editions correspond to a same person using the name as well as the dates of birth and death. This simplifies considerably the annotation.

Using this selection criterion, we found 1498 entries describing a person with a matching entry in the three other editions forming thus quadruples. Of them, 267 had a Wikidata link as the first row in Table~\ref{tab:matcheddataset}. This row corresponds to the quintuple:
\begin{small}
\begin{verbatim}
(E1_385, E2_622, E3_416, E4_473, Q215933)
\end{verbatim}
\end{small}

Most quadruples have duplicates, for instance when an entry in E1 has matches in E2, E3, and E4, and the entry in E2, has the same matching entries. Table~\ref{tab:qid} shows the number of quadruples, distinct quadruples, and quadruples with a QID (quintuples).

\begin{table}[tb]
\centering
%\resizebox{\columnwidth}{!}{
\begin{tabular}{lrrrr}
\hline
&\textbf{Quads}& \textbf{Distinct}&\textbf{Match}&\textbf{True QID}\\
\hline
     All&  1498&514&486&80\\
     QID&267&101&94&80\\
     \hline
\end{tabular}
%}
\caption{Number of quadruples, quadruples with a QID (quintuples), distinct quadruples, correctly matched quadruples, and correct QIDs assigned to a quadruple match.}
\label{tab:qid}
\end{table}

Using these quadruples, we measured the matching precision and, for the quintuples, the linking precision and recall. We decided that a quadruple was correct if it consisted of the same person. Of the 514 distinct quadruples, 486 had four times the same person, yielding a precision of 94.6\%. In this dataset, 101 quadruples also have a QID (quintuples). For this part of the dataset, 94 quadruples describe the same person. This results in a precision of 93.1\% which is about the same as for the quadruples with no QID. 

We defined the recall of the linking task as the percentage of links the system could find from the matches. Taking all quintuples, we had 80 correct QIDs. To compute F1, we considered the whole test set and the set of correct matches. We obtained a precision of about 85\% and a recall of 16.5\%, see Table~\ref{tab:f1}.
It has long been noted that there is a trade-off between precision and recall \citep{van1979information}. We had a conservative matching procedure with a high threshold that favors precision.  

The final structured dataset containing headwords, entity types, 
cross-edition matches, and Wikidata identifiers
is publicly available.\footnote{\url{https://huggingface.co/datasets/albinandersson/nf-headword-linked}} See \citetlanguageresource{andersson2026linked}.

\begin{table}[tb]
\centering
\resizebox{\columnwidth}{!}{
\begin{tabular}{lrrr}
\hline
&\textbf{Precision}&\textbf{Recall}&\textbf{F1}\\
\hline
Distinct quads&79.21&15.56&26.02\\
Correct matches&85.11&16.46&27.59\\
\hline
\end{tabular}}
\caption{Link evaluations considering all distinct quadruples and only the correct matches.}
\label{tab:f1}
\end{table}

\section{Discussion}

Table~\ref{tab:extracted_entires_res} shows there is a significant difference in the extraction results between E1-E2 and E3-E4: around 20\% roughly. We explain this with the structure of the training set, which mainly contains entries from the first two editions. The high extraction percentage of the first edition (88\%) could be due to an overfit. However, the results might stem from factors other than the simple fact that most \verb=<b>= tag occurrences are retrieved from E1-E2: 
\begin{itemize}
    \item The evolution of the encyclopedia across editions, where the patterns to define headwords have changed. Thus the same entry in two different editions may have some information discarded that was previously deemed important. The same logic applies to differences in spelling and general writing methods. Therefore, the resulting predictions may potentially differ for slightly varying entries from two different editions, mostly E1 and E4. 
    \item Even if an entry retains most information and structure across editions, it is still prone to a varying model prediction in the last two editions. This is mainly due to their low proofreading rate. 
\end{itemize}

The matching results show how people and locations covered in the encyclopedia changed between editions. Each new edition removed and added a considerable number of entries. This suggests editors actively chose which entries to include based on what was relevant at the time, rather than just adding to previous content. 

Some entries do appear across all editions from 1 to 4, likely representing people and places that remained historically important throughout this period. This kind of analysis could help us understand how encyclopedias reflect what society considered important at different times.

\section{Limitations and Future Work}
We used a semi-automatic labeling to build the training set of headwords and segmented entries. The initial rule posits that headwords are marked with \verb=<b>= tags in E1 and E2. Unfortunately, it creates a few false negatives. This can be even more confusing when two identical entries are marked differently. Cascading this problem over the rest of our 300 thousand entries in the dataset creates a question of the model's reliability if trained on a large number of false negatives. 

We would like to explore this segmentation area in future work. Should we retain the automatic construction of the dataset, we would like to lower the number of false negatives or evaluate to which degree the model is affected by potential confusion.

The headword extractor is a sequence annotator that  considers only one word. Moreover, we limited the entry categories to three types. We could include more types such as \textit{ORG} and \textit{TME}. We would also like to recover the complete annotation of the entries. This would enable us to enrich the definitions with more information such as mentions of dates, persons, and locations. We could thus extend the linking step and create relations either internally between mentions and entries of \textit{Nordisk familjebok} or externally to Wikidata.

We linked nearly 11,000 entries of \textit{Nordisk familjebok} to Wikidata. Our evaluation is nonetheless limited to a much smaller dataset. An improved evaluation would use more manually annotated data. In addition, we restricted Wikidata linking to items explicitly referencing \textit{Nordisk familjebok}. We could apply this procedure to a larger subset of the Swedish Wikipedia, or explore any relevant information contained in the linked Wikidata objects for extraction and further analysis. 

\section{Conclusion}
In this work, we described a comprehensive pipeline for processing historical encyclopedias. It consists of four major steps, notably an automated headword extraction, where we achieved an F1 score of 97.8\% and an entity type classification with an F1 score of 93.4\%. In a small-scale evaluation of the cross-edition matching, we obtained an accuracy better than 93\%. We linked approximately 11,000 Wikidata items across all editions with a precision of 85\% on our test set at the expense of recall that was of 16.5\%.

We hope that our work will offer clearer, quantifiable insights into the perspectives and viewpoints of people living during the time the encyclopedias were published. We processed the four editions and we created tools to match entries. This should improve their comparison and the understanding of how the editors and readers perceived the evolution of the world and society. Finally, it should contribute to the digital preservation of historical knowledge, making this resource more accessible for future research and analysis. Beyond the methodological contributions, this work releases three publicly 
available datasets that may support future research in historical NLP, entity linking, and digital humanities.

\section{Ethics Statement}
The collection of \textit{Nordisk familjebok} editions is in the public domain. Our work contributes to the development of tools for language resources and their annotation. We hope it can improve the understanding of human knowledge transmission through the extraction of versions of biographies and locations. Nonetheless, 
\begin{enumerate}
\item The corpus we used contains dated and possibly false information. This can notably be the case for scientific theories or technological developments.
\item The Swedish historical context and ideas of years 1870-1950 may convey biases and old-fashioned
viewpoints, possibly offensive. Users must be informed of this context.
\end{enumerate}

\section{Acknowledgements}
This work was partially supported by \textit{Vetenskaprådet}, the Swedish Research Council, registration number 2021-04533.
%\nocite{*}
\section{References}\label{sec:reference}

\bibliographystyle{lrec2026-natbib}
\bibliography{references}

\section{Language Resource References}
\label{lr:ref}
\bibliographystylelanguageresource{lrec2026-natbib}
\bibliographylanguageresource{languageresource}

\end{document}